\relax
\documentclass[letterpaper]{article} % DO NOT CHANGE THIS
\usepackage{caption}
\usepackage{subcaption}
\usepackage{booktabs}
\usepackage{aaai22}  % DO NOT CHANGE THIS
\usepackage{times}  % DO NOT CHANGE THIS
\usepackage{helvet}  % DO NOT CHANGE THIS
\usepackage{courier}  % DO NOT CHANGE THIS
\usepackage[hyphens]{url}  % DO NOT CHANGE THIS
\usepackage{graphicx} % DO NOT CHANGE THIS
\usepackage{multirow}
\usepackage{tabularx}
\usepackage[export]{adjustbox}
\usepackage{amssymb}% http://ctan.org/pkg/amssymb
\usepackage{pifont}% http://ctan.org/pkg/pifont
\usepackage{array}
\usepackage{xcolor}
\usepackage{xspace}
\usepackage{amsmath}
\usepackage{natbib}  % DO NOT CHANGE THIS AND DO NOT ADD ANY OPTIONS TO IT
\usepackage{caption} % DO NOT CHANGE THIS AND DO NOT ADD ANY OPTIONS TO IT
\DeclareCaptionStyle{ruled}{labelfont=normalfont,labelsep=colon,strut=off} % DO NOT CHANGE THIS
\usepackage{algorithm}
\usepackage{algorithmic}

\usepackage{newfloat}
\usepackage{listings}
\floatstyle{ruled}
\newfloat{listing}{tb}{lst}{}
\floatname{listing}{Listing}

\definecolor{colone}{RGB}{133, 0, 3}
\newcommand{\colone}[1]{\textcolor{colone}{#1\xspace}}
\definecolor{coltwo}{RGB}{190, 81, 7}
\newcommand{\coltwo}[1]{\textcolor{coltwo}{#1\xspace}}
\definecolor{colthree}{RGB}{13,85, 2}
\newcommand{\colthree}[1]{\textcolor{colthree}{#1\xspace}}
\definecolor{colfour}{RGB}{8, 0, 135}
\newcommand{\colfour}[1]{\textcolor{colfour}{#1\xspace}}
\nocopyright

\title{ALP: Data Augmentation using Lexicalized PCFGs for Few-Shot Text Classification}
\author{
    Hazel Kim\textsuperscript{\rm 1,2},
    Daecheol Woo\textsuperscript{\rm 1},
    Seong Joon Oh\textsuperscript{\rm 2},
    Jeong-Won Cha\textsuperscript{\rm 3},
    Yo-Sub Han\textsuperscript{\rm 1}
}
\affiliations{
    \textsuperscript{\rm 1} Yonsei University, Seoul, Republic of Korea \\
    \textsuperscript{\rm 2} NAVER AI Lab, \\
    \textsuperscript{\rm 3} Changwon National University, Changwon, Republic of Korea 
    \{hazel.kim, dwoo\}@yonsei.ac.kr, coallaoh@gmail.com,
    jcha@changwon.ac.kr,
    emmous@yonsei.ac.kr
}

\begin{document}

\maketitle

\begin{abstract}
Data augmentation has been an important ingredient for boosting performances of learned models.
Prior data augmentation methods for few-shot text classification have led to great performance boosts.
However, they have not been designed to capture the intricate compositional structure of natural language.
As a result, they fail to generate samples with plausible and diverse sentence structures.
Motivated by this, we present the \textit{data \textbf{A}ugmentation using \textbf{L}exicalized \textbf{P}robabilistic context-free grammars (ALP)} that generates augmented samples with diverse syntactic structures with plausible grammar. 
The lexicalized PCFG parse trees consider both the constituents and dependencies to produce a syntactic frame that maximizes a variety of word choices in a syntactically preservable manner without specific domain experts. 
Experiments on few-shot text classification tasks demonstrate that ALP enhances many state-of-the-art classification methods.
As a second contribution, we delve into the train-val splitting methodologies when a data augmentation method comes into play. 
We argue empirically that the traditional splitting of training and validation sets is sub-optimal compared to our novel augmentation-based splitting strategies that further expand the training split with the same number of labeled data.
Taken together, our contributions on the data augmentation strategies yield a strong training recipe for few-shot text classification tasks.
% We present a data augmentation technique using probabilistic context-free grammars to boost performance in few-shot text classification. 
% Traditional approaches in data augmentation arbitrarily select target areas of examples for synonym swap and random insertion, leverage pre-trained language models for model-based augmentation, or create synthetic examples by softly combining input and output sequences. 
% Despite their practicality, those methods are limited in capturing compositional aspects of natural language and accordingly constrain the plausible augmentation strategies.
% Our lexicalized PCFG parse trees consider both the constituents and dependencies to produce a syntactic frame that maximizes a variety of word choices in a syntactically preservable manner without specific domain experts. 
% Experiments demonstrate its good compatibility with many existing semi-supervised learning approaches including state-of-the-art classification methods in low-resource settings.
\end{abstract}

\section{Introduction}

Labeled data are an essential ingredient in training deep models. A major challenge in practice is the cost for collecting them. Data augmentation has provided a means to enlarge the training data without resorting to additional labeling cost~\citep{shorten2019survey}.
Training a text classifier is not an exception. Researchers have proposed different ways to augment text data to expand the labeled text data.
These methods focus on diversifying the word choices while preserving the labels for the classification task. 
For example, \citet{WeiZ19} arbitrarily select target areas of examples for synonym swap and random insertion. \citet{YuDLZ00L18, KumarCC20} leverage pre-trained language models for model-based augmentation. \citet{ZhangYZ20} create synthetic examples by softly combining input and output sequences. 

The prior augmentation methods have successfully diversified the samples and improved classification performances with a small number of labeled data. However, we point out a crucial shortcoming shared by those methods: they do not take into account the intricate compositional structure in natural language. Replacing words and modifying structures of a sentence without linguistic rules and guidance are likely to alter the syntax and semantics.
It can be seen in Table \ref{tab:teaser-table} that those methods (EDA, BT, and SSMBA) fail to generate samples with plausible and diverse sentence structures.

\begin{table}[t!]
    \centering
    \renewcommand{\arraystretch}{.8}
    \setlength{\tabcolsep}{.4em}
    \begin{tabular}{>{\footnotesize}c|>{\scriptsize}l}
        \toprule
        Method & \textbf{Original}: The characters didn't seem to fit very well with the book.\\
        \midrule
        \multirow{2}{3.5em}{\centering EDA}  & the characters didnt \colone{book} to fit very well with the \colone{seem}. \\
        & \colone{very} characters seem to fit the \colone{well} with the book. \\
        \midrule
        \multirow{2}{3.5em}{\centering BT} & The characters didn't seem to fit the book \colone{that well}. \\
        & The characters didn't seem to fit the book \colone{very well}. \\
        \midrule
        \multirow{2}{3.5em}{\centering SSMBA} & the \colone{the don's} seem to fit very well with the. \\
        & the characters \colone{of the also} seem to \colone{be be up} with the book. \\
        \midrule
        \multirow{2}{3.5em}{\centering ALP (Ours)} & The \coltwo{scripts} didn't \colthree{get together} very \colfour{comfortably} with the \coltwo{outlook}. \\
        & The \coltwo{prospect} didn't \colthree{satisfy} very \colfour{advantageously}. \\
        \bottomrule
    \end{tabular}
    \caption{\small \textbf{Augmented samples.} ALP generates the most diverse, syntactically plausible, and label-preserving augmented samples. EDA: Easy Data Augmenatation, BT: Back-translation, and SSMBA: Self-Supervised Manifold based Data Augmentation.}
    \label{tab:teaser-table}
\end{table}

These limitations motivate us to design a grammar-based augmentation method, which generates more plausible augmented data that better respects the syntax. We present ALP: data \textbf{A}ugmentation using \textbf{L}exicalized \textbf{P}robabilistic context-free grammars for few-shot text classification. We use lexicalized PCFG~(or L-PCFG) parse trees to consider both constituents and dependencies to capture two very different views of syntax in text data and produce a syntactic frame that maximizes a variety of word choices in a syntactically preservable manner without specific domain experts.

Our approach aims to reach theoretical guarantees of increasing both the amount and the diversity of a given dataset in a pretty label-preserving manner.
As such, ALP is designed to produce augmented samples with diverse sentence structures, each still respecting the linguistic rules and preserving the corresponding class label. The ALP samples in Table \ref{tab:teaser-table} exemplify such augmented data. We demonstrate the empirical superiority of ALP augmentation in the few-shot text classification benchmarks.

% The decent syntactic fidelity and high text diversity of ALP play a positive role in modifying labeled data for semi-supervised learning while breaking conventional beliefs that noise injection in labeled data may harm performance for consistency training~\citep{XieDHL020, ChenYY20, ChenTRBY21}. 
%Traditionally, semi-supervised learning requires a small number of clean samples as labeled data while leveraging noisy samples as unlabeled data to generalize the model better to unseen data.~\citep{ChenLCZ19, LiSH20}. 
%Data augmentation has typically applied to unlabeled data, not to labeled data~\citep{XieDHL020, ChenYY20}. 
%Contrary to common wisdom, we find that the noise in a small amount of labeled data works well in semi-supervised learning on par with that of noisy unlabeled data when labeled data is extremely limited. 
% ALP, on the other hand, works best in semi-supervised learning even though it applies augmentation operations to labeled data. Our extensive evaluation on various text classification benchmarks demonstrate its good compatibility with many existing semi-supervised learning approaches.

Recognizing the importance of the amount of data for few-shot learning tasks, we further contribute novel train-val splitting strategies that are relevant when data augmentation methods come into play. The train-val split is often regarded as a fixed constraint for a learning problem. However, we argue that the train-val split itself could be regarded as part of the model development pipeline. This viewpoint is echoed by researchers in meta-learning~\citep{setlur2020support,saunshi2021representation,BaiCZZLKWX21}, where they even suggest unconventional splitting methods like ``train-train'' that trains and validates the models on the \textit{identical} data split. We note that, when data augmentation step is part of the game, there are even further creative possibilities to split the training and validation sets. For example, under the same number of labeled data $\mathcal{S}$ and a fixed augmentation budget, we show that training on the entire augmented labeled data $\text{aug}(\mathcal{S})$ and validating on the original data $\mathcal{S}$ brings about further gains in performance across the board. 

In summary, we contribute (1) a grammar-based data augmentation method that diversifies sentence structures and (2) novel train-val splitting strategies that can be combined with general data augmentation methods.
% \begin{itemize}
%     % \item First, we show that noisy labeled data improves performance in semi-supervised learning when labeled data is very limited (i.e. 5 or 10 examples per class).
%     \item a grammar-based data augmentation method that diversifies sentence structures;
%     \item novel train-val splitting strategies that can be combined with general data augmentation methods.
% \end{itemize}

\section{Background}
In this section, we explain background research work on topics related to our core contributions: semi-supervised learning, data augmentation, and the train-val split.

\subsection{Semi-Supervised Learning (SSL)}
SSL leverages both labeled and unlabeled data for learning a discriminative task.
Early work has embraced the viewpoint that SSL is most useful when a large amount of noisy, unlabeled data source is accessible on top of a small number of clean, labeled samples~\citep{ChenLCZ19, LiSH20}. 
As such, under the SSL setup, prior studies have focused on applying data augmentation on the unlabeled data, rather than on the labeled ones. 
\citet{XieDHL020} and \citet{ChenYY20} achieve state-of-the-art model performance from noisy unlabeled data using advanced data augmentation methods in text classification with limited data. 
Contrary to the common wisdom, we find that data augmentation on the clean, labeled data also aids the generalization. Under the setup with extremely limited data ($k$-shot learning), the benefits of the increased amount of data outweighs the additional noise introduced by the augmentation algorithm.

\subsection{Data Augmentation}
Data augmentation refers to a general training technique for machine learning where the original training data are expanded to a larger set, without resorting to external sources of further data~\citep{shorten2019survey}.
Data augmentation has proven highly effective especially for deep learning and bigger models, as they generally benefit from greater amounts of data.
Data augmentation has proved useful both for labeled and unlabeled samples. For unlabeled samples, as typically done in SSL, data augmentation is applied in the form of consistency regularization~\citep{XieDHL020, ChenYY20, ChenTRBY21}. 

As mentioned above, this work focuses on the data augmentation on labeled samples. For labeled samples, the augmentation algorithm aims to preserve the semantics of the samples, while enhancing their diversity. 
Prior approaches on data augmentation for labeled text samples are limited in that they fail to observe linguistic rules and syntax.
For example, they randomly select target areas for synonym swap and random insertion \citep{WeiZ19} or leverage pre-trained language models for model-based augmentation~\citep{YuDLZ00L18, KumarCC20}. \citep{ZhangYZ20} create synthetic examples by softly combining input and output sequences. While they improve the model generalization, 
replacing words in a sentence without linguistic rules and guidance is likely to generate samples that are less realistic and plausible.
These limitations have motivated us to design our grammar-based augmentation method. We demonstrate the enhanced preservation of both the semantic and syntactic information in samples.

\begin{figure*}[t!]
    \centering
    \includegraphics[width=\linewidth]{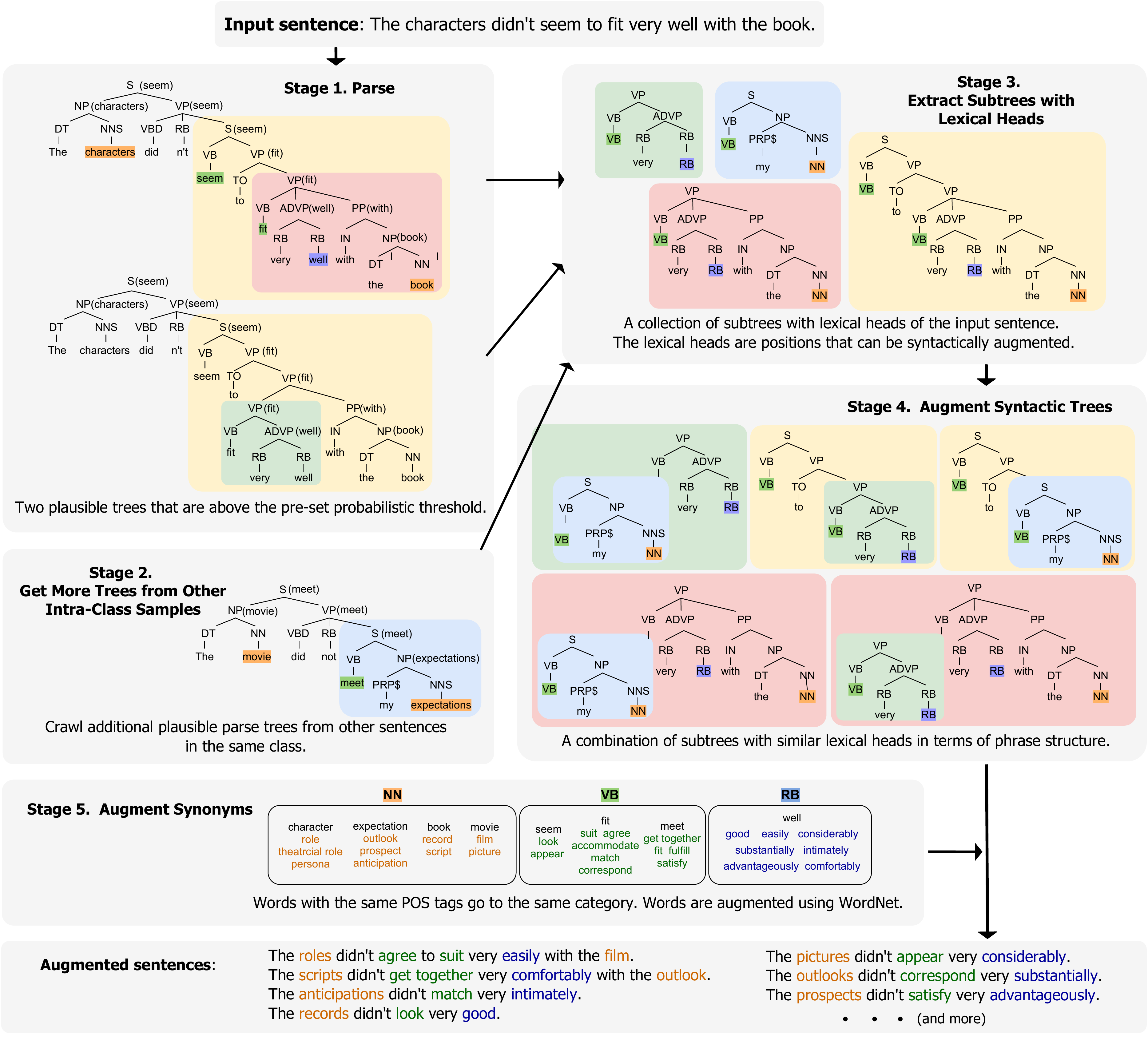}
    \caption{\small\textbf{Overview of ALP.} The step-by-step algorithm for generating augmented sentences using lexicalized PCFGs.}
    \label{fig:alp}
\end{figure*}

\subsection{Train-Val Split}

Machine learning focuses on the generalization beyond the particular training samples. Thus, a suitable segregation of data according to their dedicated uses is crucial in developing models and evaluating their generalization capabilities~\citep{hastie01statisticallearning}. Practitioners typically introduce a three-way split: train, validation, and test. Train split is used for fitting model parameters, for example via gradient descent for deep models. The validation split is used for the outer optimization problem, where hyperparameters controlling the generalization performance are fitted through black-box optimization algorithms~\citep{feurer2019hyperparameter} or heuristics~\citep{gencoglu2019hark}. The test split is the ultimate test ground for the model; the discussion of the test set is out of the scope.

The train-val split is often considered a given condition in machine learning dataset and literature. However, from a practical point of view, the ultimate crude material for building a model is the set of labeled data, which comes before the protocol for splitting it into the train and validation splits. In other words, the very protocol for the train-val split shall also be part of the overall pipeline for model building and be subject to scientific studies and solution-seeking. This view is shared by researchers in meta-learning \citep{setlur2020support,saunshi2021representation}. \cite{BaiCZZLKWX21} have even questioned the need for the train-val split and have proposed to use the entire labeled data for both training the parameters and validating the hyperparameters (the ``train-train'' method). In this work, we inherit this viewpoint, and consider various strategies for the train-val split. The space of possible splitting strategies is greatly expanded by the inclusion of the data augmentation stage.

\begin{figure*}
    \centering
    \includegraphics[width=.8\linewidth]{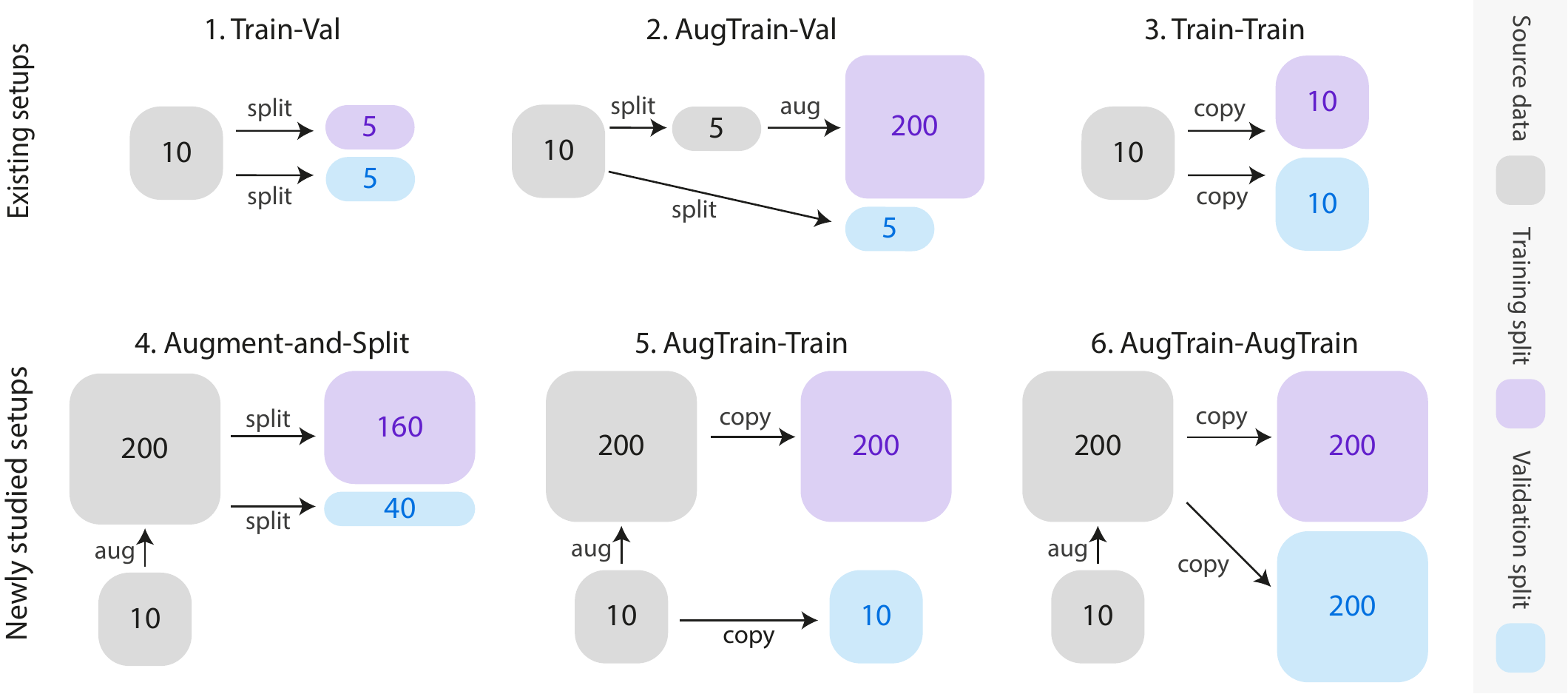}
    \caption{\small\textbf{Various train-val splitting strategies.} We match the resources used by the six splitting strategies: the number of labeled source data (10 samples per class) and the number of augmented samples (200 samples per class) if there is any. We adopt and extend the names ``train-val'' and ``train-train'' from the meta-learning literature \cite{BaiCZZLKWX21}.}
    \label{fig:setups}
\end{figure*}

\section{Data Augmentation using L-PCFGs}

Our data augmentation using lexicalized PCFGs (ALP) maximizes a variety of word choices within grammatical rules. This section introduces the ALP algorithm.
 
\subsection{Lexicalized PCFGs}

To explain the Lexicalized PCFGs, we first introduce the context-free grammar (CFG). CFG is a list of rules that define well-structured sentences in a language. Each rule has a left-hand side $\alpha$ that identifies a syntactic category, and a right-hand side $\beta$ that defines its alternative component parts. Syntactic categories include $\texttt{NP}$ for noun phrase and $\texttt{VP}$ for verb phrase.

Probabilistic context-free grammars (PCFGs) have been an important probabilistic approach to syntactic analysis~\citep{LariKYS90, JelinekFLJR92}. It assigns a probability  ${q(\alpha \to \beta)}$ to each parse tree $\alpha \to \beta$ allowed by the underlying CFGs. The parameter  ${q(\alpha \to \beta)}$ is the conditional probability of choosing rule ${\alpha \to \beta}$, given that the $\alpha$ is on the left-hand-side of the rule~\citep{Collins2013}. Under a particular type of ambiguity such as a preprositional-phrase (PP) attachment ambiguity, the PCFG model chooses a single parse tree between the two that have identical rules, depending on the value of $q(\texttt{VP} \to \texttt{VP PP})$ and $q(\texttt{NP} \to \texttt{NP PP})$. The probabilistic parser chooses a tree with $\texttt{VP} \to \texttt{VP PP}$ if $q(\texttt{VP} \to \texttt{VP PP})> q(\texttt{NP} \to \texttt{NP PP})$. The probabilistic component is crucial in our application because we aim to generate a diverse set of perturbations of a sentence based on multiple plausible hypotheses.

Lexicalized PCFGs (L-PCFGs) extends PCFGs by incorporating lexical information to further disambiguate the parsing decisions. For example, L-PCFGs enrich the graph \texttt{S $\to$ NP VP} into \texttt{S(bored) $\to$ NP(movie) VP(bored)}. The L-PCFG model is sensitive to lexical information but becomes robust because the lexical heads with the grammar rules extend the number of parameters and smooth the estimates in the model. Lexical information serves as the additional criteria to produce parse trees that are valid in the corresponding grammar rules.

\subsection{Data Augmentation using L-PCFGs}

We propose to use the rule probabilities and lexical information to diversify the grammatical choices from the limited resources. We extract many plausible subtrees using probabilistic threshold and consider lexical heads as the position information to swap and augment the syntactic structure. We substitute synonymous words within the syntactic frame.

\subsubsection{Stages 1--2. Parse with probabilistic threshold to select more trees} 
We first extract all the valid parse trees using probabilistic threshold $\tau$, instead of picking a single tree with the maximum probability. As shown in Figure~\ref{fig:alp}, the input sentence generates two valid trees that include a PP attachment ambiguity. Unlike how regular PCFGs behave, ALP picks both trees with  $\texttt{VP} \to \texttt{VP PP}$ and  $\texttt{NP} \to \texttt{NP PP}$ if  $q(\texttt{VP} \to \texttt{VP PP}) > \tau $ and  $q(\texttt{NP} \to \texttt{NP PP}) > \tau$. We use all the plausible trees generated from sentences in the same class if they are available. We expect to maximize the candidate trees to use them in the syntactic augmentation stage.

\subsubsection{Stage 3. Extract subtrees with lexical heads}  
After collecting all the plausible tree rules to use, we extract subtrees using lexical heads as the position information to swap. Figure~\ref{fig:alp} shows an example of \texttt{VP} as the lexical head. ALP swaps sub-subtrees with other types of lexical heads such as \texttt{NP} or \texttt{PP} within the subtrees if available. 

\subsubsection{Stages 4--5. Augment and Generate}  
While ALP extracts grammar rules from a starting input sentence to terminal rules, its augmentation procedure starts from bottom to top. We gather words with the same POS tags such as \texttt{NN} or \texttt{VB} into one pool as shown in Stage 5 of Figure~\ref{fig:alp}. We combine all the available words from different sentences in the same class to augment as many samples as possible, using the WordNet synonyms. The collected words have the freedom to be replaced with other words in the same POS-tagged pool. We then fill in augmented syntactic trees generated from Stage 4. The augmented syntactic phrases now replace subtrees extracted with the lexical information within the original sentence. This way, ALP preserves the label compatibility while augmenting data in the greatest number of ways.

\section{Train-Val Split with Augmented Data}

As explained in the Background section, the train-val split is crucial for ensuring good generalization performance of machine learning models. While the training and validation splits are often considered the given protocols for the sake of fair comparison among methods, the splitting of training and validation sets itself can be regarded as part of the model development framework~\citep{hastie01statisticallearning,saunshi2021representation}. Our second contribution is based on this perspective.

In this section, we delve into different possibilities of assigning the training and validation splits of a labeled source dataset, in particular in the presence of a data augmentation procedure. The motivation for searching over multiple splitting strategies is the same as that for applying data augmentation: enlarging the labeled training data for the model.

We provide conceptual diagrams for possible train-val splitting strategies based on the common labeled source dataset in Figure \ref{fig:setups}. Experiments on those strategies will be presented in the Experiments section.

\subsubsection{1--3. Existing splitting strategies}

The upper row in Figure \ref{fig:setups} corresponds to existing train-val splitting. One may split the source into disjoint train and validation splits (1. Train-Val) or may additionally apply a data augmentation algorithm on the training split (2. AugTrain-Val). In the meta-learning field, it has been argued that using the source dataset both for parameter tuning and model selection may enhance the final performance (3. Train-Train)~\citep{BaiCZZLKWX21}. The intuition is that, for low-data regime like few-shot learning, the importance of enlarging the training split outweighs the importance of segregating the validation data.

\subsubsection{4. Augment-and-Split}
The data augmentation step opens up new search spaces for the splitting strategy. For example, one may first augment the source dataset and then split the augmented data according to the ratio that ensures the maximal size of the training split while allowing for a ``just-right'' amount of validation samples. One may control the ratio between the training and validation splits (e.g. 80:20) to find the right balance. 

\subsubsection{5. AugTrain-Train}
To ensure the purity and representativeness of the validation split, one may opt for keeping the source data for validating models, while using the augmented version of the entire dataset as the training split. This ensures a large number of training data as well as minimal noise for the validation split. The cost to pay here is the overlap between the training and test splits, which may hinder the model selection based on the generalizability. However, again, under the low-data regime, the enlarged training split may bring greater gain than the loss incurred by the lack of ability to select models that generalize well.

\subsubsection{6. AugTrain-AugTrain}
In the extreme case, one may apply the ``Train-Train'' strategy on the augmented source data. Both the training and validation splits \textit{are} the augmented source data. This setup additionally enlarges the validation split to perfectly overlap with the training split. This setup is meant as a sanity check that introducing noise on the validation split via data augmentation may hinder the optimal model selection and degrade the overall performance.

\section{Experiments}

In this section, we present experimental results on our contributions. We first show the superiority of ALP among recent data augmentation baselines utilized in semi-supervised learning (SSL) for few-shot text classification. We then present various train-val splitting strategies and propose an optimal strategy that appropriately combines data augmentation with the train-val splitting.

\subsection{Experimental Setup}
Our experiments investigate three different data augmentation methods other than ALP. We measure their performances on top of three state-of-the-art SSL methods on four benchmark text classification tasks. We explain the details of those experiments here.

\subsubsection{Few-Shot Text Classification}
Usual few-shot learning refers to the setup where $k$ samples per class are available for training and other disjoint $k$ samples per class are available for validation, where $k$ is usually small ($k$-shot learning).
This means, in total, there are $2k$ labeled samples are available for each class.
We train models under the SSL fashion, by additionally utilizing the remaining data as the \textit{unlabeled} source.
% and use the remaining as unlabeled data in semi-supervised learning.
In our experiments, we consider applying data augmentation methods on the $2k$ labeled samples.
For more strategic ways to combine data augmentation with the splitting strategies for the $2k$ labeled samples, see the Section ``Train-Val Split with Augmented Data'' and Figure \ref{fig:setups}.
Unless specified otherwise, we use the ``Train-Val'' and ``AugTrain-Val'' schemes for vanilla training and the data-augmented versions, respectively.
% After augmenting the labeled data, we fine-tune the base model on newly generated data from the labeled training set to start with. 
We have conducted experiments with 5 random samplings of the labeled data, shuffling of data being presented to the models, and the weight initialization. We report the mean and standard deviation.

\subsubsection{Datasets}
We conduct experiments on four benchmark text classification tasks as summarized in Table~\ref{dataset:stats}. SST-2~\citep{socher2013} and  IMDB~\citep{MaasDPHNP11} are used for sentiment classification for movie reviews but with different sequence lengths per sample. AG News~\citep{ZhangZL15} and Yahoo~\citep{ChangRRS08} are used for topic classification in regards to news articles and question and answer pairs from the Yahoo! Answers website, respectively.

\begin{table}
\centering
\footnotesize
\renewcommand{\arraystretch}{1.2}
\begin{tabular}{ c c c c c c }
\toprule
\centering Dataset  & \#\ignorespaces Classes  & \#\ignorespaces Train & \#\ignorespaces Test & Length \\
      \midrule % from booktabs package
\centering AG News &  4   & 120K  & 7.6K  & 80 \\
\centering SST-2   &  2  & 6.9K & 1.8K  & 32     \\
\centering IMDB    &  2 & 250K & 250K  & 256     \\
\centering Yahoo! Answer & 10 & 1400K & 600K &  256     \\
 \bottomrule % from booktabs package
\end{tabular}
\caption {\small\textbf{Dataset statistics.} We report the number of samples for the original datasets. For our \textit{k}-shot classification experiments, we subsample a small portion for training and validation splits; the rest is used as the unlabeled data for semi-supervised learning.}
\label{dataset:stats}
\end{table}

\begin{table*}[t!]
\centering
\footnotesize
\renewcommand{\arraystretch}{1.4}
\makeatletter
\newcommand{\thickhline}{%
    \noalign {\ifnum 0=`}\fi \hrule height 1pt
    \futurelet \reserved@a \@xhline
}
\newcolumntype{"}{@{\hskip\tabcolsep\vrule width 1pt\hskip\tabcolsep}}
\makeatother
\begin{tabular}{ m{4em} m{2em}  m{2em}   | c c c c c }
\thickhline
                         & \multicolumn{2}{c|}{Labeled data} & \multicolumn{5}{c}{Data augmentation methods} \\ 
                         & \#\ignorespaces train & \#\ignorespaces val & No augmentation & +EDA & +BT &  +SSMBA & +ALP \\ 
\thickhline
\multirow{2}{4em}{AG News} &  \centering ${5}$ & \centering ${5}$  & 77.73 {\scriptsize $\pm$ 4.91} & 78.89 {\scriptsize $\pm$ 2.64} & 78.66 {\scriptsize $\pm$ 4.47} & 78.65 {\scriptsize $\pm$ 1.90} & $\textbf{82.30}$ {\scriptsize $\pm$ 3.34}  \\
                           &   \centering ${10}$  & \centering ${10}$ & 82.13 {\scriptsize $\pm$ 3.99} & 80.72 {\scriptsize $\pm$ 1.61} & 	83.80 {\scriptsize $\pm$ 3.48} & 84.68 {\scriptsize $\pm$ 1.07} & $\textbf{86.18}$ {\scriptsize $\pm$ 1.27}  \\
\hline
\multirow{2}{4em}{SST-2} &  \centering ${5}$ & \centering ${5}$ & 54.38 {\scriptsize $\pm$ 3.79} & 56.22 {\scriptsize $\pm$ 2.56} & 55.77 {\scriptsize $\pm$ 4.64} & 56.34 {\scriptsize $\pm$ 5.42} & $\textbf{63.40}$ {\scriptsize $\pm$ 2.33}  \\ 
                        & \centering ${10}$  & \centering ${10}$ & 61.82 {\scriptsize $\pm$ 5.85} & 53.96 {\scriptsize $\pm$ 1.40} & 62.05 {\scriptsize $\pm$ 5.03} & 59.05 {\scriptsize $\pm$ 5.70} & $\textbf{69.72}$ {\scriptsize $\pm$ 2.56} \\ 
\hline
\multirow{2}{4em}{IMDB} & \centering ${5}$ & \centering ${5}$ & 54.75 {\scriptsize $\pm$ 3.01} & 60.32 {\scriptsize $\pm$ 8.38} & 65.33 {\scriptsize $\pm$ 6.54} & 	66.43 {\scriptsize $\pm$ 9.10} & $\textbf{67.05}$ {\scriptsize $\pm$ 10.29}  \\  
                        & \centering ${10}$  & \centering ${10}$ & 68.49 {\scriptsize $\pm$ 7.42} & 69.80 {\scriptsize $\pm$ 5.75} &  70.41 {\scriptsize $\pm$ 8.96}  & 63.36 {\scriptsize $\pm$ 6.07} & \textbf{71.29 }{\scriptsize $\pm$ 6.08}   \\
\hline
\multirow{2}{4em}{Yahoo!} & \centering ${5}$ & \centering ${5}$ & 47.77 {\scriptsize $\pm$ 0.77} & \textbf{55.49} {\scriptsize $\pm$ 3.82} & 54.59 {\scriptsize $\pm$ 3.68} & 53.17 {\scriptsize $\pm$ 7.15} & 55.19 {\scriptsize $\pm$ 3.64}  \\  
                        & \centering ${10}$  & \centering ${10}$ & 58.81 {\scriptsize $\pm$  3.02} & 63.12 {\scriptsize $\pm$ 2.61} & 59.35 {\scriptsize $\pm$ 3.24} & 61.50 {\scriptsize $\pm$ 0.48} & $\textbf{64.16}$ {\scriptsize$\pm$ 1.40}  \\ 

\thickhline
\end{tabular}
\caption{\small\textbf{Comparison of data augmentation methods.} We use the Self-Training (ST) semi-supervised learning setup with $k$-shot samples for both training and validation, where $k\in\{5,10\}$. The Train-Val and AugTrain-Val splits in Figure \ref{fig:setups} have been used for No-augmentation and augmented variants, respectively.
}
\label{main_table}
\end{table*}

\subsubsection{Baseline Augmentation Methods}
We consider three data augmentation methods as our baselines. Easy Data Augmentation (EDA)~\citep{WeiZ19} is a heuristic method that randomly replaces, inserts, swaps, and deletes words. We use the official code with the recommended insertion, deletion, and swap ratios the authors provided. Unsupervised Data Augmentation~\citep{XieDHL020}, or back-translation (BT), is another common method that translates data to and from a pivot language to generate paraphrases.
We select German as intermediate languages for back-translation using FairSeq and set 0.9 as the random sampling temperature. Self-Supervised Manifold Based Data Augmentation (SSMBA)~\citep{NgCG20} generates pseudo-labels by using pre-trained masked language models as a denoising auto-encoder. SSMBA uses the corruption and reconstruction function to fill in the masked portion and thus augment the data. We use the default masked proportion and the pre-trained weights provided by the authors. 
Throughout the experiments we generate 200 samples for all augmentation methods, unless specified differently.

\subsubsection{Base Semi-Supervised Learning (SSL) Approaches}
We introduce three state-of-the-art SSL approaches to explore their compatibility with data augmentation techniques. Self-training (ST)~\cite{Yarowsky95} is a classic SSL approach using the teacher-student mechanism~\cite{Yarowsky95}. The base teacher model trained on labeled data trains the student model on unlabeled data to prevent overfitting and generalize well to unseen data. Uncertainty-aware self-training (UST)~\cite{MukherjeeA20} uses stochastic dropouts and selectively samples unlabeled examples to train the model. MixText~\cite{ChenYY20} is another SSL method using a novel data augmentation method TMix that creates virtual training samples by linearly interpolating pairs of labeled samples over their hidden-layer embeddings. MixText guesses labels for unlabeled data and leverage TMix on both labeled and unlabeled data. We use ST as the base SSL method.

%It adopts Bayesian active learning by disagreement~\citep{Houlsby:11} using stochastic dropouts to selectively sample unlabeled examples that maximize the information gain between predictions and the model posterior. UST~\citep{mukherjee2020uncertainty} minimizes the model variance not only by focusing more on unlabeled examples difficult to predict but by measuring the predictive variance to selectively focus on pseudo-labeled examples that the teacher is more confident on.

\begin{table}[t!]
\centering
\scriptsize
\renewcommand{\arraystretch}{1.2}
\setlength{\tabcolsep}{.6em}
\makeatletter
\newcommand{\thickhline}{%
    \noalign {\ifnum 0=`}\fi \hrule height 1pt
    \futurelet \reserved@a \@xhline
}
\newcolumntype{"}{@{\hskip\tabcolsep\vrule width 1pt\hskip\tabcolsep}}
\makeatother
\begin{tabular}{ l | c c | c c | c c | c c }
\thickhline
     & \multicolumn{2}{c|}{AG News} 
                        & \multicolumn{2}{c|}{SST-2} & \multicolumn{2}{c|}{IMDB} & \multicolumn{2}{c}{Yahoo!} \\ 
    Methods & 5 & 10 & 5 & 10 & 5 & 10 & 5 & 10 \\ 
\thickhline
UST  & 79.65 & 83.85 & 57.13 & 62.71 & 63.60 & 73.63 & 55.49 & 63.54 \\ 
\phantom{~} + BT & 81.61 & 83.43 & 57.22 & 67.76 & 67.14 & \textbf{83.21} & 61.78 & 63.91 \\ 
\phantom{~} + SSMBA & 83.05 & 86.32 & 48.76 & 57.00 & 61.05 & 66.82 & \textbf{62.81} & 63.63 \\ 
\phantom{~} + ALP & \textbf{84.72} & \textbf{87.41} & \textbf{73.22} & \textbf{78.01} & \textbf{71.32} & 76.33 & 61.20 & \textbf{66.89} \\ 
\hline
MixText  & 81.14 & 87.11 & 51.46 & 50.91 & 68.09 & 72.87 & 66.60 & 67.40\\ 
\phantom{~} + BT & 82.04 & 70.18 & 51.29 & 51.78 & 62.44 & 74.12 & 66.19 & 66.00 \\ 
\phantom{~} + SSMBA & 83.47 & 69.03 & 51.95 & 52.39 & 54.11 & 61.36 & 65.32 & 67.14 \\ 
\phantom{~} + ALP * & \textbf{83.50} & \textbf{87.72} & \textbf{52.44} & \textbf{57.06} & \textbf{84.35} & \textbf{84.16} & \textbf{67.31} & \textbf{67.81} \\ 

\thickhline
\end{tabular}
\caption{\small\textbf{ALP with the state-of-the-art SSL methods.} $k$-shot samples have been used for both training and validation, where $k\in\{5,10\}$.}
\label{ssl}
% \vspace{1.5em}
\end{table}
% \begin{table}[h!]
% \centering
% \footnotesize
% \renewcommand{\arraystretch}{1.4}
% \setlength{\tabcolsep}{.55em}
% \makeatletter
% \newcommand{\thickhline}{%
%     \noalign {\ifnum 0=`}\fi \hrule height 1pt
%     \futurelet \reserved@a \@xhline
% }
% \newcolumntype{"}{@{\hskip\tabcolsep\vrule width 1pt\hskip\tabcolsep}}
% \makeatother

\subsection{Evaluating ALP}

We evaluate the performance of ALP augmentation against existing methods. We then explain the performance boost in terms of the exceptionally high degrees of diversity for ALP-augmented samples.

\begin{table}[t!]
\centering
\scriptsize
\renewcommand{\arraystretch}{1.2}
\setlength{\tabcolsep}{0.3em}
\makeatletter
\newcommand{\thickhline}{%
    \noalign {\ifnum 0=`}\fi \hrule height 1pt
    \futurelet \reserved@a \@xhline
}
\newcolumntype{"}{@{\hskip\tabcolsep\vrule width 1pt\hskip\tabcolsep}}
\makeatother
\begin{tabular}{ >{\fontsize{9}{9}\selectfont}l | c c | c c c c | m{4em} }
\thickhline
    \multirow{2}{8em}{    Splitting schemes} & \multicolumn{2}{c|}{\centering \#\ignorespaces Samples} & \multicolumn{4}{c|}{Dataset} & \multirow{2}{4em}{\centering  Average} \\ 
                        & train & val & AGNews & SST-2 & IMDB & Yahoo! & \\ 
\thickhline
1. Train-Val  &  5 &  5 & 77.73 & 54.38 & 54.75 & 47.77 & \hspace{2mm} 	58.66 \\ 
2. AugTrain-Val & 200 & 5 & 82.30 & 63.40 & 64.89 & 55.19 & \hspace{2mm} 66.45 \\
3. Train-Train &  10 &  10 & 80.41 & 57.45 & 60.48 & 50.76 & \hspace{2mm} 62.28 \\ 
\hline         
4. A-and-S (50:50)   & 100 & 100 & 80.51 & \textbf{72.33} & 57.70 & 54.59 & \hspace{2mm} 66.28 \\ 
4. A-and-S (80:20) & 160 & 40 & 83.23 & 71.14 & 71.03 &  59.44 & \hspace{2mm} 72.13 \\ 
4. A-and-S (90:10) & 180 & 20 & 78.59 & 56.01 & 66.27 &60.28 & \hspace{2mm} 65.29 \\ 
5. AugTrain-Train & 200 & 10 & \textbf{83.45} & 63.64 & \textbf{80.41} & \textbf{62.82} & \hspace{2mm}  \textbf{72.40}\\ 
6. AugTrain-AugTrain & 200 & 200 & 82.53 & 64.14 & 62.53 & 59.64 & \hspace{2mm} 67.21 \\   
\thickhline
\end{tabular}
\caption{\small\textbf{Comparison of train-val splitting strategies.} See Figure \ref{fig:setups} for the description of each method. We match the resources used by the six splitting methods: the number of labeled source data (10 samples per class) and the number of augmented samples (200 samples per class) if there is any. We use ALP for the data augmentation. ``A-and-S'' refers to the Augment-and-Split scheme.
}
% \vspace{1em}
\label{DA_method}
\end{table}

\subsubsection{Comparison against Other Augmentation Methods}

Table~\ref{main_table} shows the comparison among data augmentation methods when applied to the ST semi-supervision method. We observe that the prior data augmentation methods tend to enhance the performances, with a few critical exceptions. For example, for SST-2 dataset with $k=10$, EDA drops the performance from 61.82\% to 53.96\%. Such aberrations occur at least once for EDA, BT, and SSMBA among the benchmarks considered. On the other hand, ALP \textit{uniformly improves} the performance on augmented data across the board. Moreover, ALP outperforms all previous data augmentation methods by quite a margin in general. For example, on SST-2 with $k=10$, ALP achieves 69.72\%, compared to the second-best method BT with 62.05\%.

\subsubsection{Compatibility with Various SSL Approaches}
We verify that ALP is applicable to any deep learning model by showing its performance with other semi-supervised learning (SSL) approaches, such as UST and MixText. See Table \ref{ssl} for the results. We observe that ALP generally improves the classification accuracy for both UST and MixText. 
% We observed that ALP enhanced  UST by 7\%p and MixText by 5\%p. 
We note that BT and SSMBA often fails to improved the performance for MixText; for IMDB $k=5$, they drop the accuracy by 5.44\%p and 6.73\%p, respectively. 
% Since MixText uses interpolation-based augmentation TMix for labeled and unlabeled data, we expect that the BT and SSMBA do not provide meaningful noise on labeled data for MixText while ALP does.

\begin{table}[t!]
% \vspace{-1em}
\centering
\footnotesize
\renewcommand{\arraystretch}{1.5}
\setlength{\tabcolsep}{.28em}
\makeatletter
\newcommand{\thickhline}{%
    \noalign {\ifnum 0=`}\fi \hrule height 1pt
    \futurelet \reserved@a \@xhline
}
\newcolumntype{"}{@{\hskip\tabcolsep\vrule width 1pt\hskip\tabcolsep}}
\makeatother
\begin{tabular}{ c | c c c c }
\thickhline
     & BT & SSMBA & ALP & Original \\ 
\thickhline
    \small Fidelity & 79.3{\scriptsize$\pm$4.8} & 77.6{\scriptsize$\pm$4.8} & 81.8{\scriptsize$\pm$4.6} & 85.2{\scriptsize$\pm$3.8} \\
    \hline
    \small SelfBLEU-\{2/5\} & 0.75 / 0.49 & 0.54 / 0.32 & 0.33 / 0.08 & 0.00 / 0.00 \\
\thickhline
\end{tabular}

% \footnotesize
% \begin{tabular}{ l | c c | c c | c c }
% \thickhline
%                         & \multicolumn{2}{c|}{BT}  & \multicolumn{2}{c|}{SSMBA}  & \multicolumn{2}{c}{ALP} \\ 
%                         Methods & uni & tri & uni & tri & uni & tri \\ 
% \thickhline
% AG News & \centering 29.94 & 62.92 & 22.87 & 68.05 & \textbf{45.21} & \textbf{87.44} \\ 
% SST-2 &   \centering 35.94 & 67.74 & \textbf{44.81} & 79.85 & 44.47 & \textbf{91.00} \\ 
% IMDB & \centering 27.17 & 71.47 & 21.30 & 62.10 & \textbf{35.78} & \textbf{92.55} \\ 
% Yahoo! &   \centering \textbf{24.31} & 87.27 & 18.94 & 68.15 & 24.13 & \textbf{87.41} \\ 

% \thickhline
% \end{tabular}
\caption{\small\textbf{Semantic fidelity and text diversity.} We measure semantic fidelity using BERT classifier and text diversity using SelfBLEU-$n$ where $n\in\{2,5\}$.}\label{ttr}
\label{table:diversity}
\end{table}

\begin{table}[t!]
\centering
\footnotesize
\renewcommand{\arraystretch}{1.5}
\setlength{\tabcolsep}{.4em}
\makeatletter
\newcommand{\thickhline}{%
    \noalign {\ifnum 0=`}\fi \hrule height 1pt
    \futurelet \reserved@a \@xhline
}
\newcolumntype{"}{@{\hskip\tabcolsep\vrule width 1pt\hskip\tabcolsep}}
\makeatother
\begin{tabular}{ c | c | c c c c c }
\thickhline
    $k$ & Method & No-aug & EDA & BT & SSMBA & ALP \\ 
\thickhline
\multirow{2}{1em}{\centering 5 }  & AugTrain-Val & 58.66 & 62.73 & 63.59 & 63.65 & \textbf{66.45} \\ 
                        & AugTrain-Train & 62.28 & 64.42 & 67.40 & 68.09 & \textbf{72.40} \\
                        \hline
\multirow{2}{1em}{\centering 10}  & AugTrain-Val & 67.81 & 66.90 & 68.90 & 67.15 & \textbf{72.83} \\ 
                            & AugTrain-Train & 70.97 & 67.31 & 73.15 & 68.91 & \textbf{74.89} \\ 
\thickhline
\end{tabular}
\caption{\small\textbf{Data augmentation with the best train-val split.} See Figure~\ref{fig:setups} for an overview of splitting methods. 
}
\label{best_DA_method}
\end{table}

\subsubsection{Semantic Fidelity and Text Diversity of ALP}
We provide an explanation for the superiority of ALP against baseline methods in terms of the semantic fidelity and text diversity in augmented samples.
ALP generates sentences with decent semantic fidelity, as shown in Table~\ref{table:diversity}. We use a BERT-Base classifier fine-tuned on all available labeled samples to measure classification accuracies on generated data~\citep{KumarCC20}. Higher scores indicate the preservation of class labels in generated data. ALP has average fidelity scores (81.8\%) closest to the original (85.2\%) across four different benchmarks.
We evaluate text diversity for augmentation methods by measuring Self-BLEU scores~\citep{ZhuLZGZWY18} that assess how an augmented sentence resembles the original one. Table~\ref{table:diversity} shows that ALP has the lowest Self-BLEU scores, which imply high diversity of the text. 
% ALP generates sentences with far greater type token ratios compared to BT and SSMBA. 
The diversity of sentence structure and word choices while preserving the label compatibility has an instrumental role in boosting the model performances; ALP is designed for that.
% decent type token ratios using uni-grams and generates the most diverse tri-grams. We analyze that ALP has the highest text diversity since it uses syntactic structures to diversify the data and external knowledge base to augment synonymous words. 

\subsection{Train-Val Split with Data Augmentation}

We empirically test the strategies for splitting training and validation sets from a labeled source data. As mentioned earlier, the introduction of data augmentation into the pipeline results in a multiplicity of splitting strategies in Figure~\ref{fig:setups}. Which splitting scheme will yield the best gain? 

\subsubsection{Comparison of Train-Val Splitting Strategies}

We investigate methods for splitting training and validation sets while exploring ways of taking full advantage of available labels to optimize data augmentation performance on labeled data. 
To make a fair comparison, we fix the labeled source data with $2k=10$ samples. We further fix the computational overhead due to data augmentation by fixing the number of augmented samples to 200.
Figure~\ref{fig:setups} describes the setups and Table~\ref{DA_method} shows the corresponding model performances. The Train-Val split is the standard setup for $5$-shot text classification. The baseline result is 58.66\% on average across the benchmark datasets. Adding our ALP augmentation boosts the score to 66.45\% (AugTrain-Val split). As an additional baseline, we test the Train-Train split, introduced by \citep{BaiCZZLKWX21}. As argued in the paper, we observe a mild improvement in performance (62.28\%).

We now consider more creative splitting schemes. The Augment-and-Split scheme yields the average accuracy of 72.13\% for the splitting ratio 8:2, greatly outperforming the previously considered splits like AugTrain-Val and Train-Train.
The best average performance is reported by the AugTrain-Train split (72.40\%), that uses the augmented source data for training and the original source data for validation. 
The other advantage of AugTrain-Train is that there is no additional hyperparameter attached, unlike the Augment-and-Split scheme.

We identify two lessons from the experiments here. First, few-shot classification generally benefits from an increased size of training data. This is so important that it even outweighs the importance of information segregation between training and validation. Second, for the validation set, the cleanliness often matters more than its bulk. For example, blindly increasing its size through augmentation drops the performance (5.19\%p drop from AugTrain-Train to AugTrain-AugTrain).

\subsubsection{Compatibility of AugTrain-Train with Various Augmentation Methods}

We now turn to the question: does our best splitting strategy, AugTrain-Train, also yield the best results for data augmentation methods other than ALP? Table~\ref{best_DA_method} shows the results.
We observe that our novel split, AugTrain-Train, uniformly improves the performance for EDA, BT, and SSMBA for $k\in\{5,10\}$. This validates the effectiveness of AugTrain-Train beyond ALP. We further confirm that under this new split, ALP is still the best-performing augmentation method.
% hat train-train method improves performances with other data augmentation methods while ALP still outperforms all the data augmentation baselines.

% \subsection{Limitations}

\section{Conclusions}

We have introduced a novel text augmentation method, ALP, that considers the syntactic structure of the augmented samples. Using grammar-based mechanisms ALP increases the diversity of the sentence structures and the word choices in sentences, while preserving the semantic content. With the exceptionally high level of diversity, ALP outperforms existing text augmentation methods on the few-shot text classification tasks on four real-world benchmarks. Given the importance of securing a large amount of labeled training data, we also explore novel train-val splitting schemes for few-shot classification task. We show that the usual disjoint training and validation splits are in fact sub-optimal and propose a novel scheme that uses the augmented source data as the training split and the un-augmented original source as validation. The two contributions are orthogonal. Together, they comprise a powerful recipe for greatly enhancing the few-shot classification scores across the board.

%\newpage
\section{Acknowledgements}
This research was supported by the NRF grant
(NRF-2020R1A4A3079947), IITP grant (No.~2021-0-00354) and 
the AI Graduate School Program (No.~2020-0-01361) funded by 
the Korea government~(MSIT).
Han is a corresponding author.

\bibliography{aaai22}

\end{document}